\documentclass[10pt,twocolumn,letterpaper]{article}

\usepackage{wacv}
\usepackage{times}
\usepackage{epsfig}
\usepackage{graphicx}
\usepackage{amsmath}
\usepackage{amssymb}
\usepackage{booktabs}
\usepackage{multirow}

\def\wacvPaperID{1428} 

\wacvalgorithmstrack   

\wacvfinalcopy 

\ifwacvfinal
\usepackage[breaklinks=true,bookmarks=false]{hyperref}
\else
\usepackage[pagebackref=true,breaklinks=true,colorlinks,bookmarks=false]{hyperref}
\fi

\begin{document}

\title{High-Resolution Depth Estimation for 360$^{\circ}$ Panoramas through \\ Perspective and Panoramic Depth Images Registration}

\author{Chi-Han Peng\\
National Yang Ming Chiao Tung University\\
{\tt\small pengchihan@nycu.edu.tw}
\and
Jiayao Zhang\\
ByteDance\\
{\tt\small jiayao.zhang@bytedance.com}
}

\maketitle
\thispagestyle{empty}

\begin{abstract}

We propose a novel approach to compute high-resolution (2048x1024 and higher) depths for panoramas that is significantly faster and qualitatively and qualitatively more accurate than the current state-of-the-art method~\cite{reyarea2021360monodepth}. As traditional neural network-based methods have limitations in the output image sizes (up to 1024x512) due to GPU memory constraints, both~\cite{reyarea2021360monodepth} and our method rely on stitching multiple perspective disparity or depth images to come out a unified panoramic depth map. However, to achieve globally consistent stitching,~\cite{reyarea2021360monodepth} relied on solving extensive disparity map alignment and Poisson-based blending problems, leading to high computation time. Instead, we propose to use an existing panoramic depth map (computed in real-time by any panorama-based method) as the common target for the individual perspective depth maps to register to. This key idea made producing globally consistent stitching results from a straightforward task. Our experiments show that our method generates qualitatively better results than existing panorama-based methods, and further outperforms them quantitatively on datasets unseen by these methods.
\end{abstract}

\section{Introduction}

Panoramas with depth information are very useful for 3D computer vision tasks such as novel view synthesis~\cite{Kopf2020,OmniPhotos,xu2021layout}, 3D scene understanding (\eg room layout estimation~\cite{Sun_2019_CVPR}), omnidirectional SLAM~\cite{9196695}, and virtual reality (VR) applications~\cite{Serrano_TVCG_VR-6dof}. Traditional monocular depth estimation methods built for perspective images cannot directly work on panoramas, as a panorama cannot be converted to a perspective image since the latter can not have field-of-view (FOV) angles exceeding 180 degrees. Therefore, many depth estimation methods that are specially built for panoramas have been proposed~\cite{zioulis2018omnidepth,wang2020bifuse,sun2021hohonet,pintoreslicenet,jiang2021unifuse,li2021panodepth,Li2022CVPR}.

Most of these methods are based on deep neural networks trained on panorama datasets with ground-truth depths such as Matterport3D~\cite{chang2017matterport3d} and Stanford2D3D~\cite{2017arXiv170201105A}. The ground-truth depths in such datasets have been calibrated such that they are {\em globally consistent} (i.e., having the same scale and shift) across all the viewing directions from the camera position. As a result, these methods estimate consistent depth maps for the whole panoramas.

\begin{figure}[t]
  \centering
  \includegraphics[width=1\linewidth]{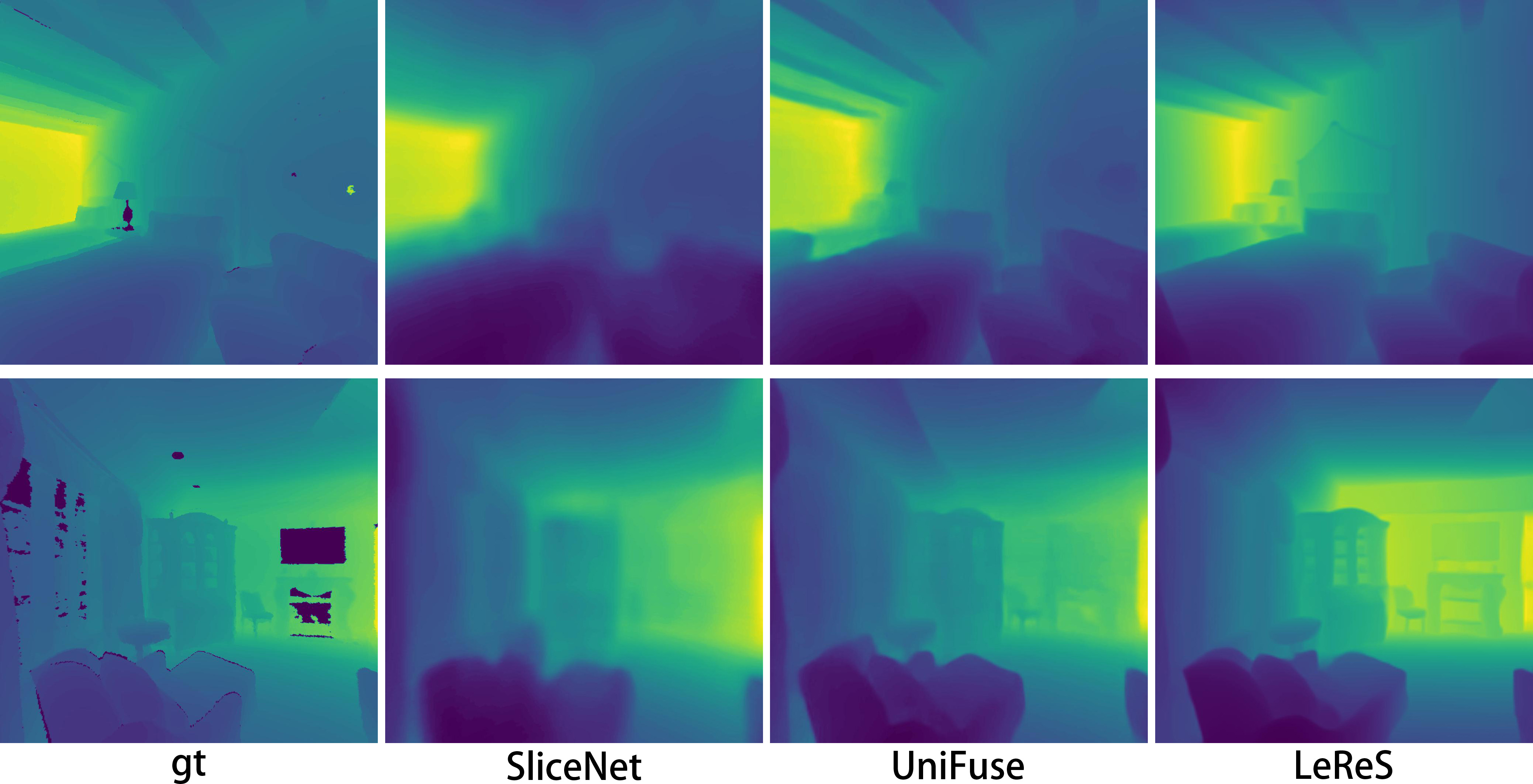}
  \caption{\label{fig:pmaps} Comparisons of depth estimations by panorama-based methods (SliceNet~\cite{pintoreslicenet} and UniFuse~\cite{jiang2021unifuse}), and perspective method (LeRes~\cite{Wei2021CVPR}) of the same region in a panorama. We find that LeReS produced qualitatively better results with better levels of details than any panorama-based methods by a large margin.} 
\end{figure}

However, we observe two main shortcomings of these methods. First, due to the limitation of GPU memory during the training of the CNN architecture being used, they can only output depth images with resolutions as high as 1024x512. This is far below the native resolutions of RGB panoramas (most modern commodity 360$^{\circ}$ cameras can shoot panoramas with resolutions exceeding 4096x2048) and is insufficient for VR applications. For example, a 90$^{\circ}$-by-90$^{\circ}$ perspective view rendered out of a 1024x512 panorama would have a resolution of only 256x256. Second, we observe that depth maps estimated by panorama-based methods often lack the same level of detail as the results generated by perspective methods. This may be because they were trained on panoramic RGB-D datasets, which are much smaller and less diverse than the traditional perspective RGB-D datasets that perspective methods were trained on. See Figure~\ref{fig:pmaps} for examples.


To address the shortcomings, the authors of~\cite{reyarea2021360monodepth} proposed a {\em stitching}-based approach as follows. First, the panorama is partitioned into several perspective views. They used a 20-view partition mimicking the 20 triangle faces of an icosahedron (\ie "tangent images" in~\cite{Eder_2020_CVPR}). Each view is fed to a modern perspective depth estimation method to get a perspective disparity map (they used MiDaS~\cite{9178977}). Each perspective map is projected back to a common equirectangular domain. A fully panoramic depth map can then be computed by stitching together the projected perspective disparity maps and then converted to depth values. The main challenge of such stitching-based methods is that the perspective maps tend to have different scales and shifts of values because the perspective depth estimation method was run on different fields of view of the panorama. To fix this inconsistency problem, they relied on extensive disparity map alignment and Poisson blending strategies, both become very expensive for high-resolution images.

Instead, we propose a simple solution to the global consistency problem: {\em leveraging a panoramic depth map as the common target for the perspective depth maps to register to}. Note that such "reference" panoramic depth maps can be generated in real-time by an existing panorama-based method, which is known to produce globally consistent values. Our stitching pipeline is as follows.

We begin with partitioning the panorama into several rectangular regions. For each region, we generate a perspective depth map using LeReS~\cite{Wei2021CVPR}. Next, we generate a full panoramic depth map using a recent panorama-based method such as SliceNet~\cite{pintoreslicenet} or Unifuse~\cite{jiang2021unifuse}. For each perspective map, we register a low degree (\eg quadratic or cubic) transform function by fitting a set of sampled pixels in the perspective map to the corresponding pixels in the panoramic map in a least-squared sense. Being an optimization problem with only a few variables, the computational cost of this step is very low.

The previous step minimized the differences of scales and shifts between the perspective maps but does not completely erase the visible seams in between. Therefore, we further blend the registered perspective maps by a Poisson-based approach similar to~\cite{10.1145/882262.882269}. We opt for computing the final panoramic depth map as the one that fits the Laplacians of the perspective depth maps in a least-squared sense. As the optimization problem is ill-posed (being shift invariant), we regularize it by the $L2$ distances to the reference panoramic depth map with a small weight. Note that our blending approach is similar to the disparity map blending step in~\cite{reyarea2021360monodepth} with the main difference that our approach no longer needs the spatial-varying weights (\eg "radial" or "frustum"-shaped) in~\cite{reyarea2021360monodepth} to enforce smoother transitions between the perspective maps.



Through experiments, we show that our method produced qualitatively and qualitatively better results than~\cite{reyarea2021360monodepth} at a significantly faster speed. Our contributions are summarized as follows:

\begin{itemize}
\item We propose a simple, effective, and cost-efficient solution to the global consistency problem of stitching-based methods for high-resolution panoramic depth map generations, namely the registration-based approach. 

\item Benefited from the effectiveness of the aforementioned step, we find that a straightforward Poisson blending-based approach is sufficient to blend the registered perspective depth maps and erases the visible seams in between, without the needs for complex spatial-varying weights as was done in~\cite{reyarea2021360monodepth}.

\end{itemize}

\section{Related Work}
\label{sec:related_work}

\subsection{Panorama-based 3D Modeling and Datasets}

Leveraging the increasing popularity of 360$^{\circ}$ cameras, 3D modeling of indoor scenes based on panoramic image inputs have become a popular research field in recent years. Key tasks include depth estimation~\cite{zioulis2018omnidepth,wang2020bifuse,pintoreslicenet,jiang2021unifuse}, room layout estimation~\cite{zou2018layoutnet,yang2019dula,SunHSC19,Pintore:2020:AI3,zeng2020joint,wang2021led2}, object detection and segmentation~\cite{xia2018gibson,sun2021hohonet}, and more generally 3D reconstruction tasks such as registration of multiple panoramas~\cite{yang2020noise,Haiyan2022}.

Panoramic image datasets with depth information are summarized as follows. Matterport3D~\cite{chang2017matterport3d} and Standford2D3D~\cite{armeni2017joint} provide real-world panoramas of diverse indoor scenes with ground-truth depths. Note that although the depths of one panorama are captured through multiple 3D scans (the 3D scanners have limited field-of-views and relied on motors to take multiple scans in a controlled manner), they have been calibrated so that the depth values are consistent across all viewing directions - we don't see any seams between the different views in the depth maps. SunCG~\cite{song2016ssc}, Structure3D~\cite{Structured3D}, and Replica360~\cite{replica19arxiv} are synthetic datasets of indoor scenes. They provide photo-realistically rendered indoor RGB-D images and annotations of 3D structures including room layouts. 3D60~\cite{zioulis2019spherical} is a collective dataset consisting of real and synthetic sources.


\subsection{Monocular perspective depth estimation}



Monocular (\ie needed just a single image as input instead of stereo ones) perspective depth estimation is a very active research topic in which modern methods can now reliably predict generally accurate depths from arbitrary images in the wild (\ie in unseen datasets). We list methods proposed in recent years with competitive performances: Xian et al. 2018~\cite{Xian_2018_CVPR}, MegaDepth~\cite{MDLi18}, MiDaS v2~\cite{9711226} and v3~\cite{9178977}, SGR~\cite{Xian_2020_CVPR}, Huynh et al. 2021~\cite{Huynh2021}, AdaBins~\cite{9578024}, and LeReS~\cite{Wei2021CVPR}. Note that some of the methods predict disparity values (e.g., MiDaS), which are the inverse of depths. Authors of~\cite{Miangoleh2021Boosting} proposed a derivative method that calls an external perspective depth estimation method repeatedly on carefully chosen subsets of the input image, and blend the partial depth estimations to form the final output. The downside is that it is much slower than standard methods. 

\subsection{Panoramic Depth Estimation}

Tateno et al.~\cite{Tateno_2018_ECCV} proposed a deformable convolutional filter that maps perspective views to the equirectangular domain in a distortion-aware manner and used it to perform panoramic depth predictions via numerous, densely sampled perspective depth-prediction calls that could be trained on both panoramic and perspective datasets for perspective depth predictions. In~\cite{eder2019mapped} an elaborate discussion of such filters are discussed. OmniDepth~\cite{zioulis2018omnidepth} was an early method to estimate depths specifically for panoramas in an end-to-end manner. They reported that applying monocular perspective methods directly on equirectangular images led to inferior results. To solve the problem, they trained a customized encoder-decoder network (called "UResNet" in their paper) directly on panoramas with ground-truth depths from synthetic sources. Later, BiFuse~\cite{wang2020bifuse} found that injecting perspective views of the panoramas (in cubemap formats) into the training process increased the performance considerably. Inspired by the gravity-aligned nature of indoor scenes, SliceNet~\cite{pintoreslicenet} proposed a novel network architecture that encodes the input panorama into feature vectors that each correspond to a vertical slice of the panorama. HoHoNet~\cite{sun2021hohonet} leveraged a similar idea in their method that predicts room layouts, depths, and semantic labels simultaneously. Unifuse~\cite{jiang2021unifuse} iterates on the "two-branch" (one feeds the equirectangular projection and another feeds the cubemap projection of the input panorama) network design of BiFuse and proposed a simpler version. PanoDepth~\cite{li2021panodepth} uses stereo matching ideas to improve panoramic depth estimations. OmniFusion~\cite{Li2022CVPR} is a panoramic method via aligning and blending tangent depth maps using transformers. A trend of network design in recent methods is to leverage perspective views sampled on panoramic images while using various strategies to eliminate the distortions and discontinuity artifacts caused in the process~\cite{9428385,Li2022CVPR,shen2022panoformer}. Finally, 360MonoDepth~\cite{reyarea2021360monodepth} proposed a novel derivative approach to generate high resolution panoramic depth maps by stitching perspective depth maps generated by external methods.


\section{Method}
\label{sec:method}

\begin{figure*}[t]
  \centering
  \includegraphics[width=1\linewidth]{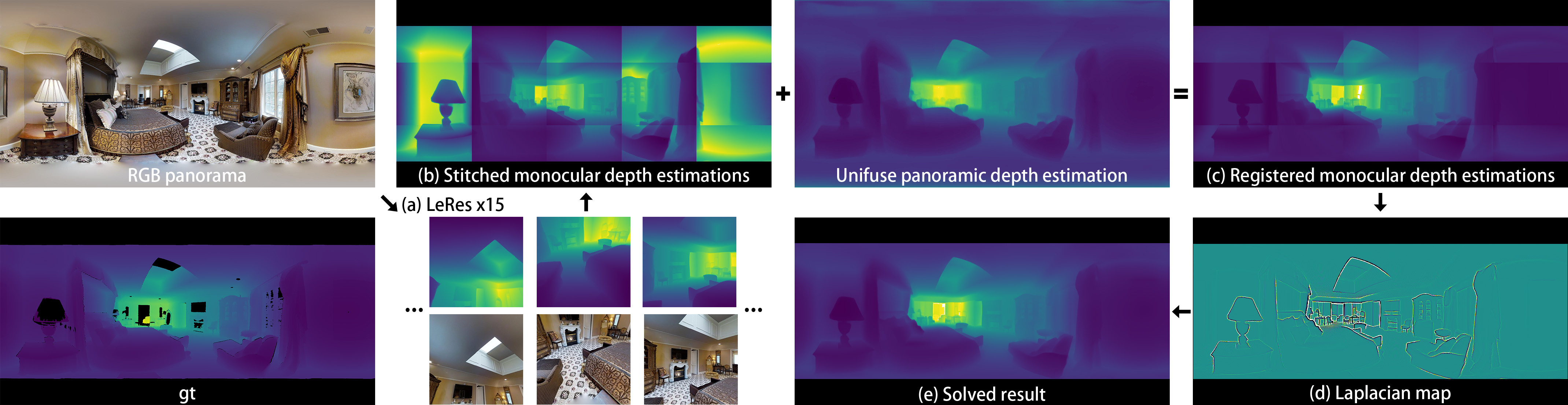}
  \caption{\label{fig:overview} Overview of our method. (a) We first partition the panorama into several perspective views and feed each perspective image to a monocular depth estimation method such as LeReS~\cite{Wei2021CVPR}. Note that most panoramic RGB-D datasets don't have depths in the top and bottom so our method skips those regions. (b) The predicted perspective depth maps are projected to the equirectangular domain and are stitched together to form a panoramic depth map. (c) For each projected perspective depth map, we solve a low-degree function in a least-squared sense that transforms the pixels in the perspective map to the corresponding pixels in a shared "reference" panoramic depth map, which is generated in real-time by a panorama-based method such as SliceNet~\cite{pintoreslicenet} or UniFuse~\cite{jiang2021unifuse}. (d) We generate a panoramic Laplacian map of the registered depth maps. (e) Finally, we optimize for a new panoramic depth map that fits the Laplacians of the registered depth maps with a small regularization term using L2 distances to the reference panoramic depth map.}
\end{figure*}

We use notations as follows. We assume a 3D world space with right-hand rule in which the +z axis is the "straight-up" direction and the +x axis is chosen arbitrarily. A {\em viewing direction} is equivalent to a point on the unit sphere centered at the origin, which can be denoted by a spherical coordinate, $(\phi,\theta)$, in which $0^{\circ} \le \phi < 360^{\circ}$ is the {\em azimuth angle} from the +x axis on the x-y plane in counter-clockwise order and $0^{\circ} \le \theta \le 180^{\circ}$ is the {\em zenith angle} to the +z axis. An {\em equirectangular projection} maps a viewing direction to a 2D point trivially as $(\phi,\theta)$. The {\em equirectangular domain} denotes the rectangle $(0^{\circ},0^{\circ})$, $(360^{\circ},0^{\circ})$, $(360^{\circ},180^{\circ})$, $(0^{\circ},180^{\circ})$ in the 2D plane. A {\em perspective view} is a perspective projection defined by a viewing direction and field-of-view angles in the horizontal and vertical directions ($FOVx$ and $FOVy$ angles in short) in which the camera is at the origin and +z axis is the "up" direction. An {\em perspective-to-equirectangular (P2E) projection} projects a perspective view to a region in the equirectangular domain. Note that the projected region would not be a rectangle anymore (see Figure~\ref{fig:P2E}). A summary of our method is given in Figure~\ref{fig:overview} and its description. We explain the key steps in the following sub-sections.

\subsection{Equirectangular-to-Perspective Partitions}
\label{sec:partition}

\begin{figure*}[t]
  \centering
  \includegraphics[width=1\linewidth]{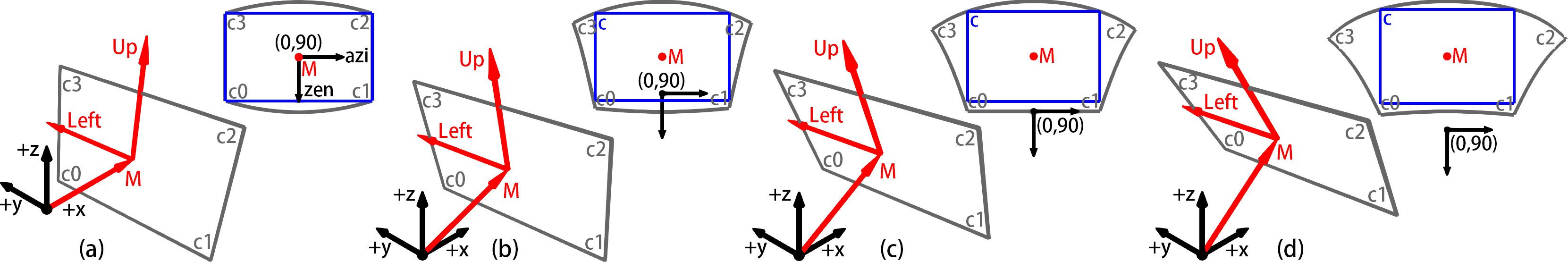}
  \caption{\label{fig:P2E} (a) to (d) show four cases of a perspective view with $FOVx = 80^{\circ}$ and $FOVy = 60^{\circ}$ in 3D space where in (a) the looking direction, $M$, 's spherical coordinate is $(0,90^{\circ})$ (on the equator), in (b) it is $(0,70^{\circ})$ (slightly tilted up), in (c) it is $(0,60^{\circ})$ (tilted up and the bottom edge of the image plane lies on the x-y plane), and in (d) it is $(0,50^{\circ})$ (further tilted up). $c0$ to $c3$ denote the four corners of the image plane. $Left$ and $Up$ denote the two axes of the image plane in 3D. In each figure's upper-right corner, we show the P2E-projected region of the perspective view in the equirectangular domain. We show the partition that each view is designated to cover in blue.} 
\end{figure*}

Our method produces depths for a rectangular subset of the equirectangular domain that spans azimuth from $0^{\circ}$ to $360^{\circ}$ and zenith from $25^{\circ}$ to $155^{\circ}$, where all panorama datasets have depth values. We denote this region as the "target domain". We partition the target domain into rectangular sub-regions along horizontal and vertical lines, and then use a perspective depth estimation method such as LeReS~\cite{Wei2021CVPR} to estimate the depth values for each sub-region. We denote a rectangular region that covers azimuth from $\phi_0^{\circ}$ to $\phi_1^{\circ}$ and zenith from $\theta_0^{\circ}$ to $\theta_1^{\circ}$ as a "partition" $P_{\phi_0,\phi_1,\theta_0,\theta_1}$. Recall that we assume all perspective views have the camera position at origin and the "up" direction at +z axis. Even so, there are still infinitely many ways to construct a perspective view that sufficiently covers a partition. Therefore, we describe an algorithm to find one such perspective view that would bound the region tightly as follows.

First we note that the shape of a perspective view's P2E-projected domain is invariant to rotations along the z-axis, so we limit our analysis to the case of azimuth equals zero. Also, there are reflective symmetries vertically w.r.t. the equator so we need only to analysis the "tilted-up" cases (zenith not greater than $90^{\circ}$) and the corresponding tilted-down cases can be derived trivially. Now, as shown in Figure~\ref{fig:P2E}, we observe all possible cases of a perspective view's P2E-projected domains and rectangles centered at the viewing direction (point $M$ in the equirectangular domain) bounded within. These rectangles are the rectangular partitions that a perspective view may cover. We find that in the equirectangular domain, the upper edges of the P2E-projected domain are always closer or equal to the viewing direction ($M$) than the lower edges along the y-axis. Therefore, we choose the rectangle with left and right edges horizontally aligned to the bottom two corners (same azimuths) of the P2E-projected domain that goes as up as possible until touching the upper edge.

The viewing direction $M$'s 3D coordinate is $(cos(\Theta),0, sin(\Theta))$, $\Theta$ is the vertical tilted-up angle from the equator. The two axis of the image plane in 3D are $Left = (0,1,0)$ and $Up = (-sin(\Theta), 0, cos(\Theta))$, respectively. To derive the lower-left corner ($c0$)'s azimuth, we first find $c0$'s 3D position as:
\begin{equation}
M + tan(FOVx/2) * Left - tan(FOVy/2) * Up, 
\end{equation}
$FOVx$ and $FOVy$ are the field-of-view angles of the perspective view. $c0$'s azimuth can then be computed and denoted as $c0_\phi$. Note that the 3D positions of the other three corners (denoted as $c1$, $c2$, and $c3$, in counterclockwise order) can be derived similarly. We can now uniquely locate the point $c$ on the upper edge with azimuth equals to $c0_\phi$.

Finally, given a rectangular partition that the perspective view is designated to cover, we derive the corresponding $c_\phi$ and $c_\theta$. $c$'s 3D position is now fixed. We then solve $FOVy$ such that the upper edge would intersect $c$. Next, we solve $FOVx$ using the formula for $c0$'s azimuth. $\blacksquare$

For each perspective view of the panorama, the corresponding monocular depth map is computed by a perspective method such as LeReS. By default, we partition the target domains into 3 rows and 5 columns (15 total) of rectangular sub-regions. The rows are divided along zeniths $25^{\circ}$, $60^{\circ}$, $120^{\circ}$, and $155^{\circ}$. The columns are divided along azimuths $0^{\circ}$, $72^{\circ}$, $144^{\circ}$, $216^{\circ}$, and $288^{\circ}$. Comparisons of different ways to partition the target domain are in Section~\ref{sec:analysis}.

\subsection{Perspective-to-Equirectangular Registration}
\label{sec:registration}

As shown in Figure~\ref{fig:overview} (b), each partition of the target domain is filled with depth values from the corresponding P2E-projected perspective depth map, using looking directions and $FOVx$ and $FOVy$ chosen according to the algorithm described in Section~\ref{sec:partition}. However, each filled partition tends to have different scales and shifts of values. With the common scale and shift being unknown, as has been shown in~\cite{reyarea2021360monodepth}, aligning them using optimization could be an expensive process, especially for high-resolution images.

Instead, we propose the idea of using a "reference" panoramic depth map, in which the depth values are known to be consistent across the whole target domain, as the common target for the perspective depth maps to register to. The reference panoramic depth map can be generated by one of the existing panorama-based methods such as SliceNet, or UniFuse. For each partition $P_{\phi_0,\phi_1,\theta_0,\theta_1}$, the registration is summarized as solving a linear least-squares optimization problem as follows:
\begin{equation}
\begin{aligned}
& \underset{a,b,c,d}{\text{argmin}}
& & \sum_{i=0}^{N-1} (a (x_i)^3 + b (x_i)^2 + c x_i + x_i - X_i)^2
\end{aligned}
\end{equation}
where $x_i$ and $X_i$ denote the depth values of the $i$-th sampled pixel in the partition and the corresponding pixel in the reference panoramic depth map, respectively. $N$ denotes the number of sampled pixels in the partition. We sample pixels at every 1 degree horizontally and vertically. Here, we show an algorithm design with a cubic function. We also experimented with other choices (\eg quadratic and linear) and report the finding in Section~\ref{sec:analysis}. We use Google Ceres Solver~\cite{Agarwal_Ceres_Solver_2022} to solve the optimization problem. We then use the solved registration function to transform every pixel values in a partition (\ie not just the sampled pixels).


\subsection{Laplacian-based Poisson Blending}
\label{sec:Laplacian}

The aforementioned registration process aligned the scales and shifts of the depth values in every partition (\eg comparing Figure~\ref{fig:overview} (b) and (c)) but does not completely remove the visible seams in between. To solve this issue, we blend the partitions using a Poisson-based approach. Similar to the traditional gradient-based Poisson blending algorithms (which was used in~\cite{reyarea2021360monodepth}), we opt for optimizing for a panoramic depth map that fits the Laplacians of the registered perspective maps directly. 


We use the standard 3x3 discrete Laplacian operator (shown in Equation~\ref{eq:Laplacian}). For each partition, we also sample "padded" depth values outside its region in the equirectangular domain by slightly expanding the equirectangular regions of each partition and updating the FOV angles respectively. By default, we expand the region of each partition 5 pixels horizontally and 2 pixels vertically (for both 2K and 4K cases). This means that there would be small overlaps of Laplacian values between the partitions and we take the average of Laplacians at the overlapping pixels. This padding step serves the purpose of slightly smoothing out the Laplacians between adjacent partitions. The Laplacian-based blending is expressed as an optimization problem as follows:
\begin{equation}
\begin{aligned}
& \underset{x_{i,j}, \: 0 \le i < W, \: 0 \le j < H}{\text{argmin}}
& & \sum_{j=1}^{H-2} \sum_{i=1}^{W-2} ( (l_{i,j} - L_{i,j})^2 \\
& & & + \gamma \, (x_{i,j} - X_{i,j})^2 ) \\
& \text{subject to}
& & x_{i,j} \ge 0, \forall i,j
\end{aligned}
\end{equation}
where $x_{i,j}$ is the depth value at pixel coordinate $(i,j)$ in the panoramic depth map to be solved. $W$ and $H$ are the width and height of the panoramic depth map. $l_{i,j}$ is the Laplacian value at pixel $(i,j)$ computed by the standard 3x3 discrete Laplacian operator:
\begin{equation}
\label{eq:Laplacian}
\begin{aligned}
l_{i,j} = 4 x_{i,j} - x_{i-1,j} - x_{i+1,j} - x_{i,j-1} - x_{i,j+1}, \\ \forall \: 1 \le i \le W-2, 1 \le j \le H-2.
\end{aligned}
\end{equation}
and $L_{i,j}$ are the "target" Laplacian values computed by the same formula in the registered depth values at each partitions (one example is shown in Figure~\ref{fig:overview} (d)). $X_{i,j}$ is the depth value at pixel coordinate $(i,j)$ in the reference panoramic map. In short, the objective function has two terms - a Laplacian term and a regularization term using L2 distances to the reference panoramic depth map. We set $\gamma$, the weight for the regularization term, to 1e-4.

The optimization problem is solved using the standard Jacobi iterative method with the depth values of the reference panoramic depth map as the initial guess. To speed up solving, we solve the problem in a multi-scale manner. That is, we first solve the problem in a reduced-resolution version of the image buffer, pass the solved values to a finer-resolution buffer, solve again, and so on until the problem is solved in the original resolution. For the 2048x1024 case, we solve in 3 levels (\ie 512x256, 1024x512, and 2048x1024). The iterations for the finest level are chosen to be 50 as we observed that the problem usually converged (with residuals lower than 0.1\% of initial values) during iteration 40 to 50. We then set iterations for the subsequent levels to be 100 and 200. For the 4096x2048 case, we solve in 4 levels (50, 100, 150, and 200 iterations).

\section{Results and Analysis}
\label{sec:results}

We tested our method and~\cite{reyarea2021360monodepth} on a computer with Intel i7-10700 CPU, 32GB ram, and NVidia RTX 2070 GPU. We compare the timing statistics of our method versus~\cite{reyarea2021360monodepth} in Table~\ref{tab:timing}. In summary, our method is 3.05 times faster (2048x1024 outputs), and 1.94 times faster (4096x2048 outputs), than~\cite{reyarea2021360monodepth}. We draw each perspective view in a 1024x989 resolution, which is enough to form 4K outputs when stitched. Same as in prior methods (including OmniDepth, BiFuse, SliceNet, and UniFuse), when comparing a computed depth map, $\omega$, to a ground-truth depth map, $\Omega$, a "median-scaling" step is first taken by multiplying every depth value in $\omega$ by the median value of $\Omega$ divided by the median value of $\omega$. Note that for the Matterport dataset, depth values are reported in meters. We choose HoHoNet~\cite{sun2021hohonet}, SliceNet~\cite{pintoreslicenet}, and UniFuse~\cite{jiang2021unifuse}, as the methods to generate the panoramic depth maps for our method.



\begin{table}
\begin{center}
\begin{tabular}{| c || c | c | c | c | c || c |}
\hline
Res & Pers.D & Pano.D & Reg. & Blend & Total & \cite{reyarea2021360monodepth} \\
\hline
2K & 4.27 & 0.122 & 0.22 & 10.89 & 15.50 & 47.3 \\
\hline
4K & 4.27 & 0.122 & 0.23 & 34.27 & 38.89 & 75.3 \\
\hline
\end{tabular}
\label{tab:timing}
\caption{Timing comparisons. We list the times (in seconds) for our perspective-view rendering and depth estimation (LeReS 15 times), panoramic depth estimation (UniFuse), registration step (cubic), and blending step. We solved for 100 panoramas and calculated the average. For~\cite{reyarea2021360monodepth}, to reproduce similar results as shown in their paper, we set the options to: Poisson blending, 3 levels of disparity map alignments, and MiDaS V2. Note that~\cite{reyarea2021360monodepth} also has a one-time "pre-processing" step which took about 40 seconds.}
\end{center}
\end{table}




\subsection{Quantitative Comparisons}
\label{sec:performance}

In Table~\ref{tab:comparisons}, we quantitatively compare our method versus recent panorama-based methods and~\cite{reyarea2021360monodepth} on the Matterport dataset testing split in 2K resolution outputs and the synthetic Replica360 dataset in 2K and 4K resolutions provided by~\cite{reyarea2021360monodepth}. Same as in~\cite{reyarea2021360monodepth}, we up-sample 1K results generated by previous methods to 2K by bilinear interpolation. Our method produced quantitatively better results than~\cite{reyarea2021360monodepth} on all three datasets by large margins. Our method, when paired with SliceNet or HoHoNet, scored nearly as good as the top performers (SliceNet and HoHoNet) on the Matterport 2K dataset. But when tested on the Replica360 2K and 4K datasets, in which the previous panorama-based methods were not trained on, our method outperforms all existing panorama-based methods. Also note that on Replica360 2K and 4K datasets, our method improved the quantitative score of the panorama-based method that is being used in every cases. We find~\cite{reyarea2021360monodepth} to perform competitively on the Replica360 2K and 4K datasets (through accuracy metrics), but not on the Matterport 2K dataset.


\begin{table}[t!]
\small
\begin{center}
\begin{tabular}{| c | c c c c|}
\hline
Method & RMSE & MAE & AbsRel & RMSE$_{log}$\\
\hline
Reg. only & +3.97\% & +8.49\% & +14.62\% & +8.23\% \\
Blend only & +8.32\% & +12.23\% & +26.69\% & +56.13\% \\
No padding & +0.02\% & +0.00\% & +0.01\% & +0.13\% \\
Avg. s/s  & +6.79\% & +11.26\% & +24.18\% & +47.20\% \\
Smoothing & +6.36\% & +9.92\% & +17.49\% & +28.01\% \\
4-fold & +0.36\% & +0.70\% & +1.51\% & +1.47\% \\
3-fold & +0.05\% & +0.41\% & +0.08\% & +0.35\% \\
MiDaS & +0.26\% & +1.06\% & +1.75\% & +2.89\% \\
Linear & -0.52\% & -0.77\% & -1.53\% & -1.60\% \\
Quadratic & -0.51\% & -0.75\% & -0.88\% & -0.77\% \\
\hline
\end{tabular}
\label{tab:ablation}
\caption{Quantitative statistics of ablation studies and alternative design choices. We show relative ratios w.r.t. the case of using UniFuse as the panorama-based method on Matterport 2K dataset. "Reg. only" means only performing the registration step. "Blend only" means only performing the blending step. "No padding" means skipping the padding in the blending step. "Avg. s/s" means simply registering each partition to the average scale and shift. "Smoothing" means replacing the blending step by smoothing depth values at partition boundaries. "4-fold" means taking a partition of 3 rows by 4 columns (horizontally divided along azimuths $0^{\circ}$, $90^{\circ}$, $180^{\circ}$, and $270^{\circ}$), and "3-fold" means taking a partition of 3 rows by 3 columns (along azimuths $0^{\circ}$, $120^{\circ}$, and $240^{\circ}$). "MiDaS" means using MiDaS for perspective depth estimations. "Linear" and "Quadratic" means using alternative degree registration functions.}
\end{center}
\end{table}

\begin{table*}
\begin{center}
\begin{tabular}{| c | c | c c c c | c c c|}
\hline
\multicolumn{2}{|c|}{} & \multicolumn{4}{c|}{Error metric $\downarrow$} & \multicolumn{3}{c|}{Accuracy metric $\uparrow$} \\
DS & Method & RMSE & MAE & AbsRel & RMSE$_{log}$ & $\delta_1$ & $\delta_2 $ & $\delta_3$ \\
\hline

\parbox[t]{2mm}{\multirow{7}{*}{\rotatebox[origin=c]{90}{Matterport 2K}}} &  Bifuse~\cite{wang2020bifuse} & 0.6350 & 0.3675 & 0.1367 & 0.0901 & 82.77\% & 94.46\% & 97.53\% \\
& HoHoNet~\cite{sun2021hohonet} & \textcolor{blue}{0.4707} & \textcolor{blue}{0.2620} & \textcolor{blue}{0.0967} & \textcolor{green}{0.0629} & \textcolor{blue}{90.50\%} & \textcolor{green}{97.27\%} & 97.09\% \\
& SliceNet~\cite{pintoreslicenet} & \textcolor{red}{0.4463} & \textcolor{red}{0.2153} & \textcolor{red}{0.0665} & \textcolor{red}{0.0513} & \textcolor{red}{95.17\%} & \textcolor{red}{98.07\%} & \textcolor{red}{99.54\%} \\
& UniFuse~\cite{jiang2021unifuse} & 0.6040 & 0.3309 & 0.1110 & 0.0728 & 87.79\% & 95.70\% & 98.38\% \\
& 360MonoDepth~\cite{reyarea2021360monodepth} & 0.7729 & 0.5106 & 0.2653 & 0.1253 & 60.38\% & 85.55\% & 94.70\% \\
& Our (HoHoNet) & \textcolor{blue}{0.4791} & 0.2655 & 0.1004 & 0.0662 & 90.23\% & 97.09\% & \textcolor{blue}{98.93\%} \\ 
& Our (SliceNet) & 0.4949 & \textcolor{green}{0.2569} & \textcolor{green}{0.0883} & \textcolor{blue}{0.0648} & \textcolor{green}{91.51\%} & \textcolor{blue}{97.21\%} & \textcolor{green}{99.10\%} \\ 
& Our (UniFuse) & 0.6107 & 0.3333 & 0.1152 & 0.0766 & 87.08\% & 95.36\% & 98.20\% \\
\hline

\parbox[t]{2mm}{\multirow{7}{*}{\rotatebox[origin=c]{90}{Replica360 2K}}} & Bifuse~\cite{wang2020bifuse} & 0.0555 & 0.0416 & 0.2150 & 0.1121 & 71.56\% & 91.39\% & 96.32.\% \\
& HoHoNet~\cite{sun2021hohonet} & \textcolor{green}{0.0300} & \textcolor{green}{0.0193} & \textcolor{green}{0.1116} & \textcolor{green}{0.0671} & \textcolor{green}{90.31\%} & 95.90\% & \textcolor{blue}{98.11\%} \\
& SliceNet~\cite{pintoreslicenet} & 0.0403 & 0.0279 & 0.1590 & 0.0896 & 85.15\% & 93.88\% & 96.44\% \\
& UniFuse~\cite{jiang2021unifuse} & 0.0362 & 0.0248 & 0.1336 & 0.0774 & 86.87\% & \textcolor{blue}{95.94\%} & 97.72\% \\
& 360MonoDepth~\cite{reyarea2021360monodepth} & 0.0706 & 0.0456 & 0.1813 & 0.0865 & 78.48\% & 93.56\% & \textcolor{red}{98.34\%} \\
& Our (HoHoNet) & \textcolor{red}{0.0272} & \textcolor{red}{0.0182} & \textcolor{red}{0.1074} & \textcolor{red}{0.0643} & \textcolor{red}{90.98\%} & \textcolor{green}{96.07\%} & \textcolor{green}{98.28\%}  \\ 
& Our (SliceNet) & 0.0380 & 0.0272 & 0.1553 & 0.0862 & 85.35\% & 94.31\% & 96.77\%  \\ 
& Our (UniFuse) & \textcolor{blue}{0.0354} & \textcolor{blue}{0.0247} & \textcolor{blue}{0.1334} & \textcolor{blue}{0.0762} & \textcolor{blue}{87.00\%} & \textcolor{red}{96.08\%} & 97.80\% \\
\hline

\parbox[t]{2mm}{\multirow{7}{*}{\rotatebox[origin=c]{90}{Replica360 4K}}} & Bifuse~\cite{wang2020bifuse} & 0.0642 & 0.0485 & 0.2446 & 0.1266 & 63.27\% & 89.13\% & 95.65\% \\
& HoHoNet~\cite{sun2021hohonet} & \textcolor{green}{0.0357} & \textcolor{green}{0.0249} & \textcolor{green}{0.1359} & \textcolor{green}{0.0744} & \textcolor{green}{85.17\%} & 94.63\% & 96.61\% \\
& SliceNet~\cite{pintoreslicenet} & 0.0473 & 0.0341 & 0.1891 & 0.0994 & 78.31\% & 93.17\% & 96.77\% \\
& UniFuse~\cite{jiang2021unifuse} & 0.0394 & 0.0289 & 0.1480 & 0.0818 & 82.20\% & \textcolor{green}{96.26\%} & \textcolor{green}{98.54\%} \\
& 360MonoDepth~\cite{reyarea2021360monodepth} & 0.0611 & 0.0400 & 0.1667 & 0.0815 & 80.04\% & \textcolor{blue}{95.25\%} & \textcolor{blue}{98.47\%} \\
& Our (HoHoNet) & \textcolor{red}{0.0332} & \textcolor{red}{0.0239} & \textcolor{red}{0.1309} & \textcolor{red}{0.0709} & \textcolor{red}{86.07\%} & 94.98\% & 96.76\% \\ 
& Our (SliceNet) & 0.0444 & 0.0335 & 0.1831 & 0.0975 & 76.80\% & 93.27\% & 97.34\% \\ 
& Our (UniFuse) & \textcolor{blue}{0.0380} & \textcolor{blue}{0.0281} & \textcolor{blue}{0.1447} & \textcolor{blue}{0.0795} & \textcolor{blue}{82.69\%} & \textcolor{red}{96.66\%} & \textcolor{red}{98.65\%} \\
\hline
\end{tabular}
\end{center}
\label{tab:comparisons}
\caption{Quantitative comparisons of our method (using HoHoNet, SliceNet, or UniFuse to generate the reference panoramic depth maps) versus previous panorama-based methods and the stitching-based method proposed in~\cite{reyarea2021360monodepth}. RMSE, MAE, AbsRel, and RMSE$_{log}$ measure the root mean squared error, mean absolute error, mean relative error, and RMSE in log-10 space (same as in UniFuse and BiFuse), of depth values. $\delta_1$, $\delta_2$, and $\delta_3$ measure the ratios of pixels with mutual relative errors below $1.25$, $1.25^2$, and $1.25^3$, respectively. Highlighting: \textcolor{red}{best}, \textcolor{green}{second-best}, \textcolor{blue}{third-best}.}
\end{table*}


\subsection{Qualitative Comparisons}
\label{sec:qualitative}

\begin{figure*}[t]
  \centering
  \includegraphics[width=1\linewidth]{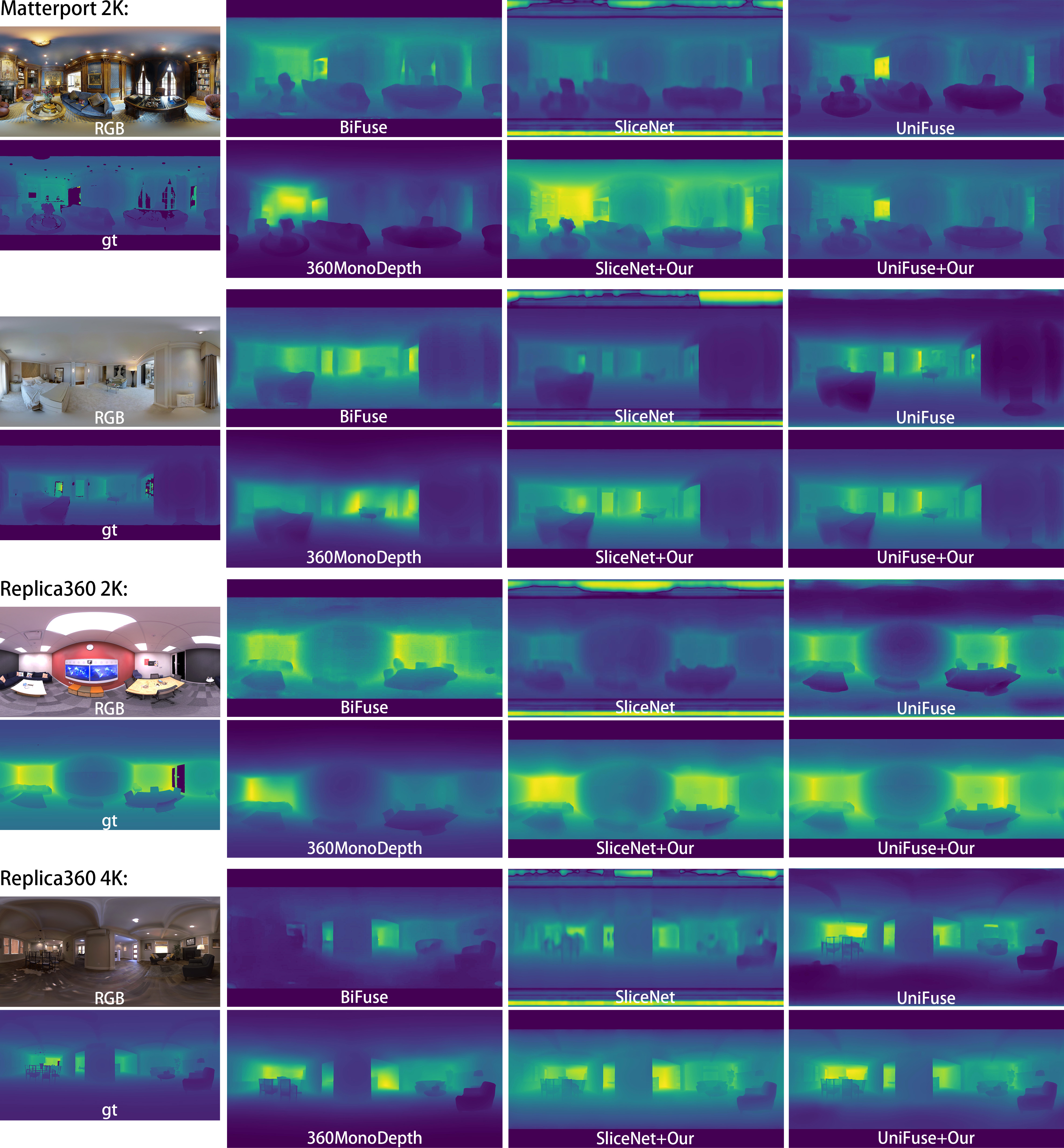}
  \caption{\label{fig:qualitative_matterport_Replica}. Qualitative comparisons of results generated by our method (using SliceNet or UniFuse to generate the reference panoramic depth maps), previous panorama-based methods, and the stitching-based method in~\cite{reyarea2021360monodepth} (360MonoDepth).} 
\end{figure*}

We show qualitative comparisons in Figure~\ref{fig:qualitative_matterport_Replica} on the Matterport 2K and Replica360 2K and 4K datasets. We found UniFuse produced qualitatively best results among panorama-based methods. For stitching-based methods, we found~\cite{reyarea2021360monodepth} generates clearer images than UniFuse, while our method generated slightly clearer results than~\cite{reyarea2021360monodepth} with finer details in general. We found that in~\cite{reyarea2021360monodepth}'s results, the estimated depths can deviate from the ground truth when observed at larger scales. For example, observe the inconsistent depths of the two white walls (at roughly the same distances to the camera) of the Replica360 2K example in Figure~\ref{fig:qualitative_matterport_Replica}. More examples are shown in the Supplementary Materials.

\begin{figure}[t]
  \centering
  \includegraphics[width=1\linewidth]{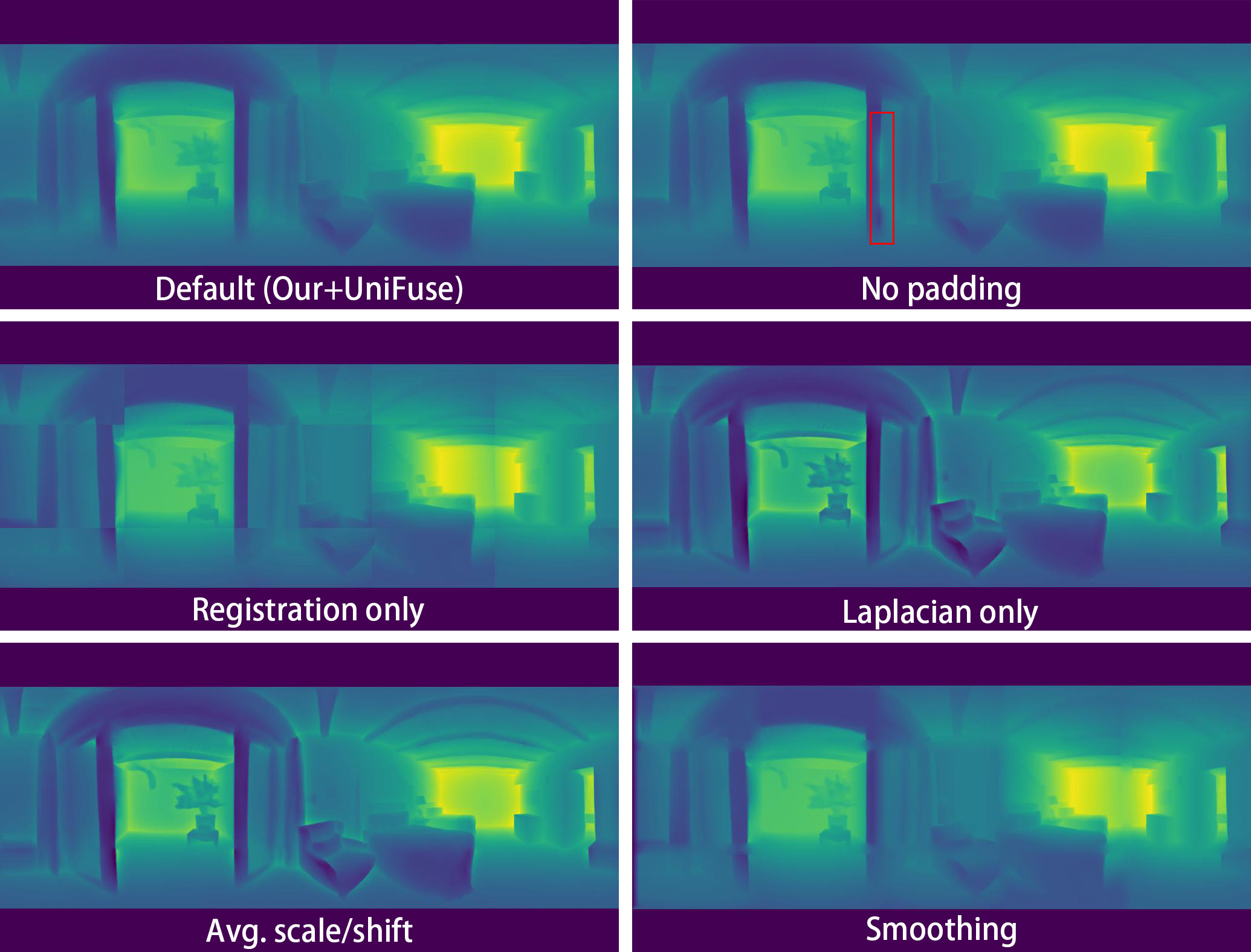}
  \caption{\label{fig:ablations}. Qualitative comparisons of ablation studies and alternative design choices. Only performing the blending step led to over-saturated results. Similarly, replacing the registration step with a trivial averaging of scales and shifts led to similar results. Skipping the padding may lead to glitches at partition boundaries. The "smoothing" result shows that a simple color smoothing cannot effectively remove the inconsistency between partitions.}
\end{figure}

\subsection{Analysis}
\label{sec:analysis}

See Table~\ref{tab:ablation} and Figure~\ref{fig:ablations} for supporting quantitative statistics and qualitative examples. 

\noindent {\bf Ablation studies:} Only performing the registration step led to blocks with visible seams (\eg Figure~\ref{fig:overview} (c)). Similarly, only performing the blending step led to worse quantitative and qualitative results. Skipping the padding in the blending step may led to glitches at the partition boundaries. Next, we tried swapping key designs of our method with trivial approaches. First, instead of using a common reference map, we tried to register the individual perspective depth maps by transforming them to the average of scale and shift. Second, we tried blending the registered perspective depth maps by simply smoothing the depth values at partition boundaries. In both cases, the results become worse quantitatively and qualitatively.

\noindent {\bf Alternative degree registration functions:} We tried linear and quadratic registration functions instead of cubic functions. We found that using simpler functions actually led to slightly better quantitative scores. We still opt for cubic functions because we think the additional degrees of freedom may be beneficial for coping with unseen cases.

\noindent {\bf Switching to MiDaS:} To verify that the differences of performance between~\cite{reyarea2021360monodepth} and our method are not simply due to the different perspective methods that were used, we tried using MiDaS instead of LeReS to do the perspective depth estimations. We find the results to be slightly worse, but do not make up all the differences between the two methods.

\noindent {\bf Different partitions:} We tried other ways to partition the target domain into perspective views. We find that sparser partitions (larger FOVs) led to slightly worse performance. This may be because perspective methods run on larger FOVs output less detailed estimations in general (see discussions in~\cite{Miangoleh2021Boosting}). A benefit of using fewer partitions is shorter computation times of the perspective depth maps generation step (which is not the performance bottleneck).


\noindent {\bf Limitations:} Our results may be negatively affected by major errors of the panoramic depth maps being used. For examples, for unseen cases such as outdoor scenes, some of the existing panorama-based methods may produce very low quality results and negatively affect our results (two such examples are OmniDepth~\cite{zioulis2018omnidepth} and BiFuse run on outdoor scenes, as shown in~\cite{reyarea2021360monodepth}'s website). Same as in~\cite{reyarea2021360monodepth}, another limitation is the accuracy of the estimated perspective depth maps being used. In either cases, our method is poised to benefit from advances in both panoramic and perspective depth estimation methods.


\section{Conclusion}
\label{sec:conclusion}

We show that a bottleneck of stitching-based panoramic depth estimation methods, \ie the global consistency problem, can be solved satisfactorily and very efficiently by our registration-based approach. Accordingly, we propose a streamlined stitching pipeline that outperforms current state-of-the-arts method~\cite{reyarea2021360monodepth} quantitatively and qualitatively and is much faster. For future work, our main goal is speed improvement, either by developing GPU-based Laplacian solvers inspired by~\cite{10.1145/1618452.1618462} to solve the blending step or alternatively, using style transfer-based approaches. \textbf{Acknowledgements:} this work is funded by the Ministry of Science and Technology of Taiwan (110N007 and 111R10286C).




{\small
\bibliographystyle{ieee_fullname}
\bibliography{egbib}
}

\clearpage
\section{Supplementary Materials}

\subsection{Additional Qualitative Results}

\begin{figure*}[t]
  \centering
  \includegraphics[width=1\linewidth]{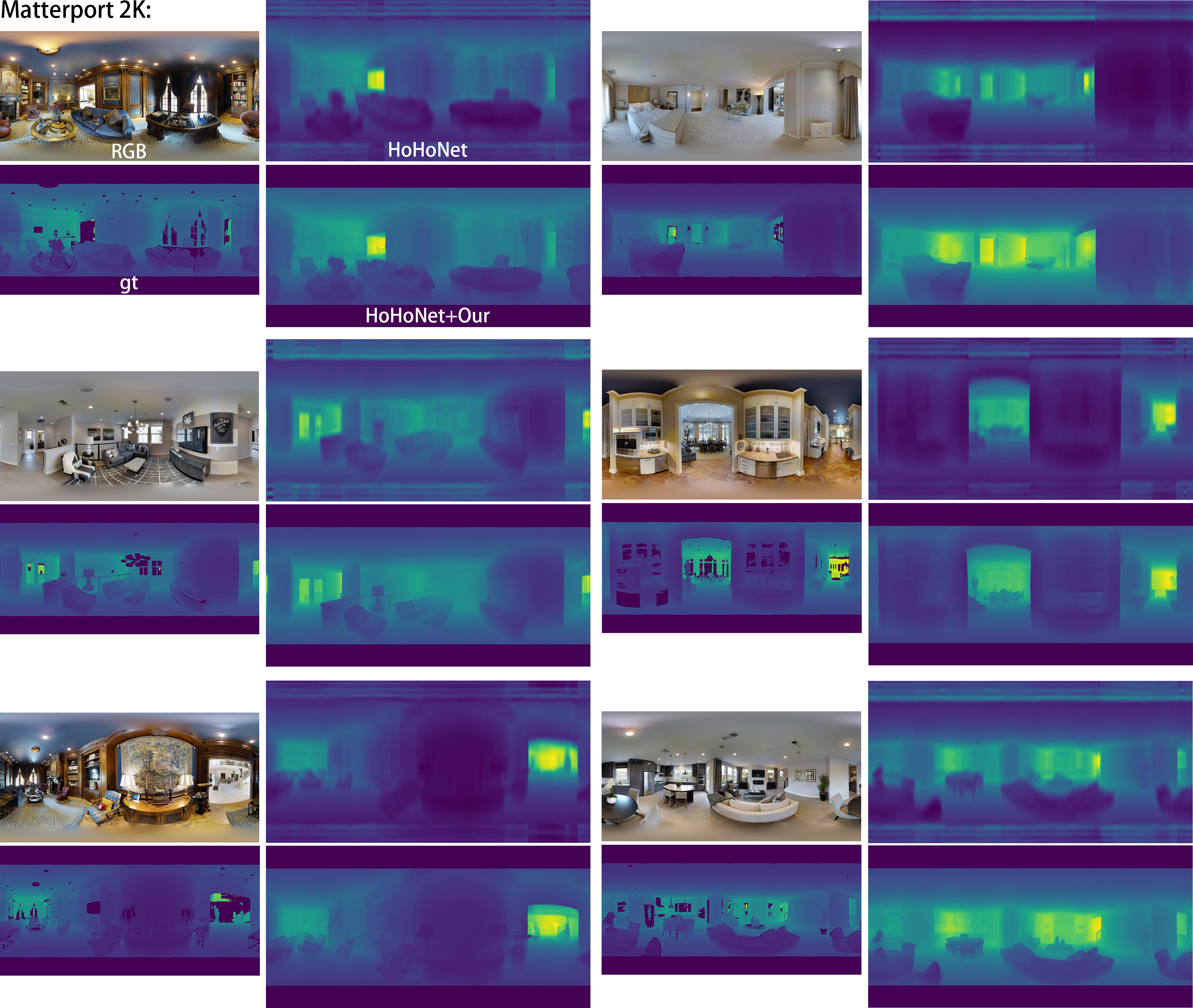}
  \caption{\label{fig:qualitative_hohonet_matterport}. Qualitative comparisons on the Matterport 2K dataset of HoHoNet and our method using HoHoNet to generate the reference panorama depth maps.} 
\end{figure*}

\begin{figure*}[t]
  \centering
  \includegraphics[width=1\linewidth]{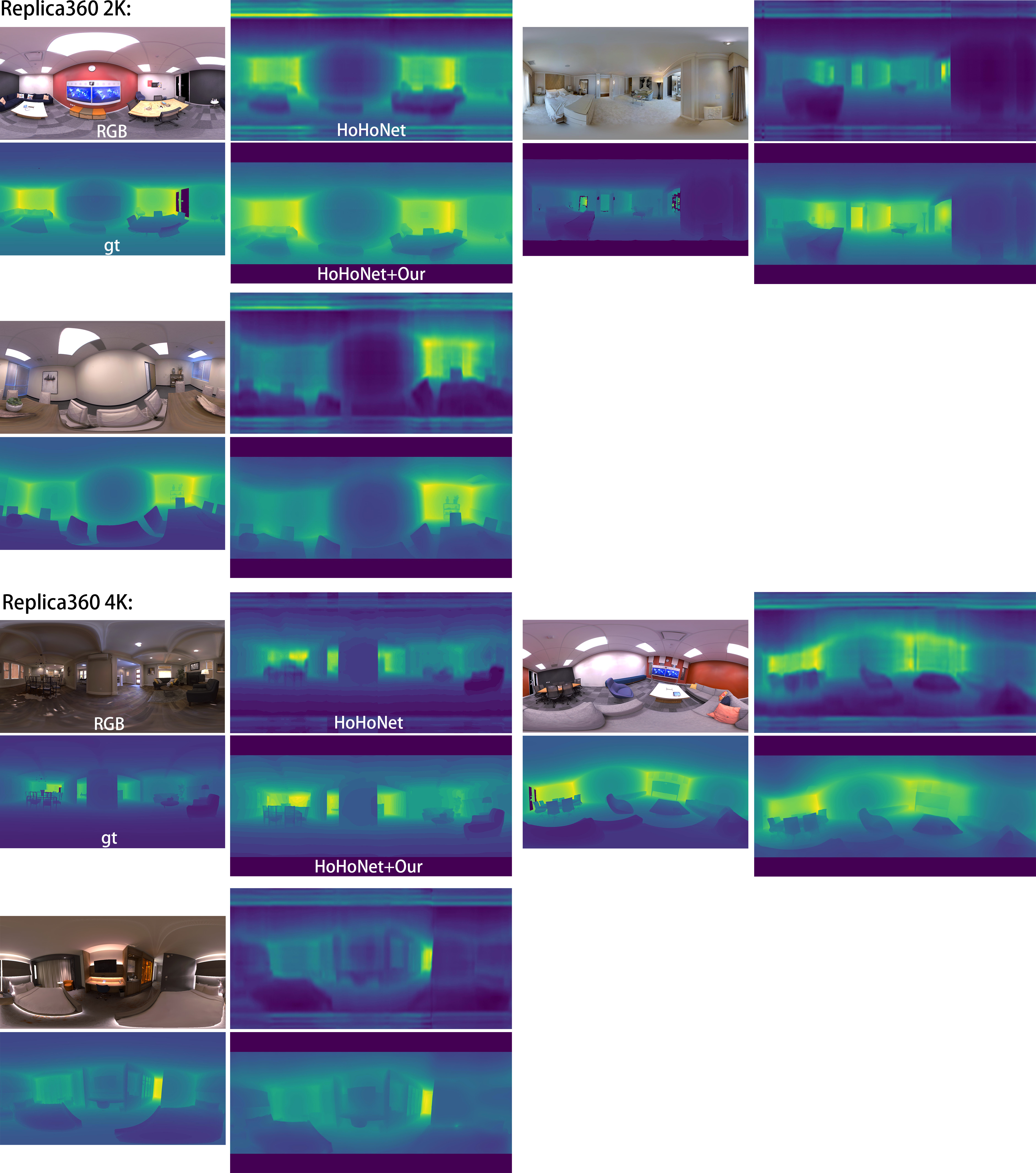}
  \caption{\label{fig:qualitative_hohonet_replica}. Qualitative comparisons on the Replica360 2K and 4K dataset of HoHoNet and our method using HoHoNet to generate the reference panorama depth maps.} 
\end{figure*}

We show qualitative comparisons of HoHoNet~\cite{chang2017matterport3d}'s results versua our method using HoHoNet to generate the reference panoramic depth maps in Figure~\ref{fig:qualitative_hohonet_matterport} and Figure~\ref{fig:qualitative_hohonet_replica}. Same as in other cases (\eg SliceNet and UniFuse), we can see that our method significantly improved the levels of details of the respective panorama-based method.

\begin{figure*}[t]
  \centering
  \includegraphics[width=1\linewidth]{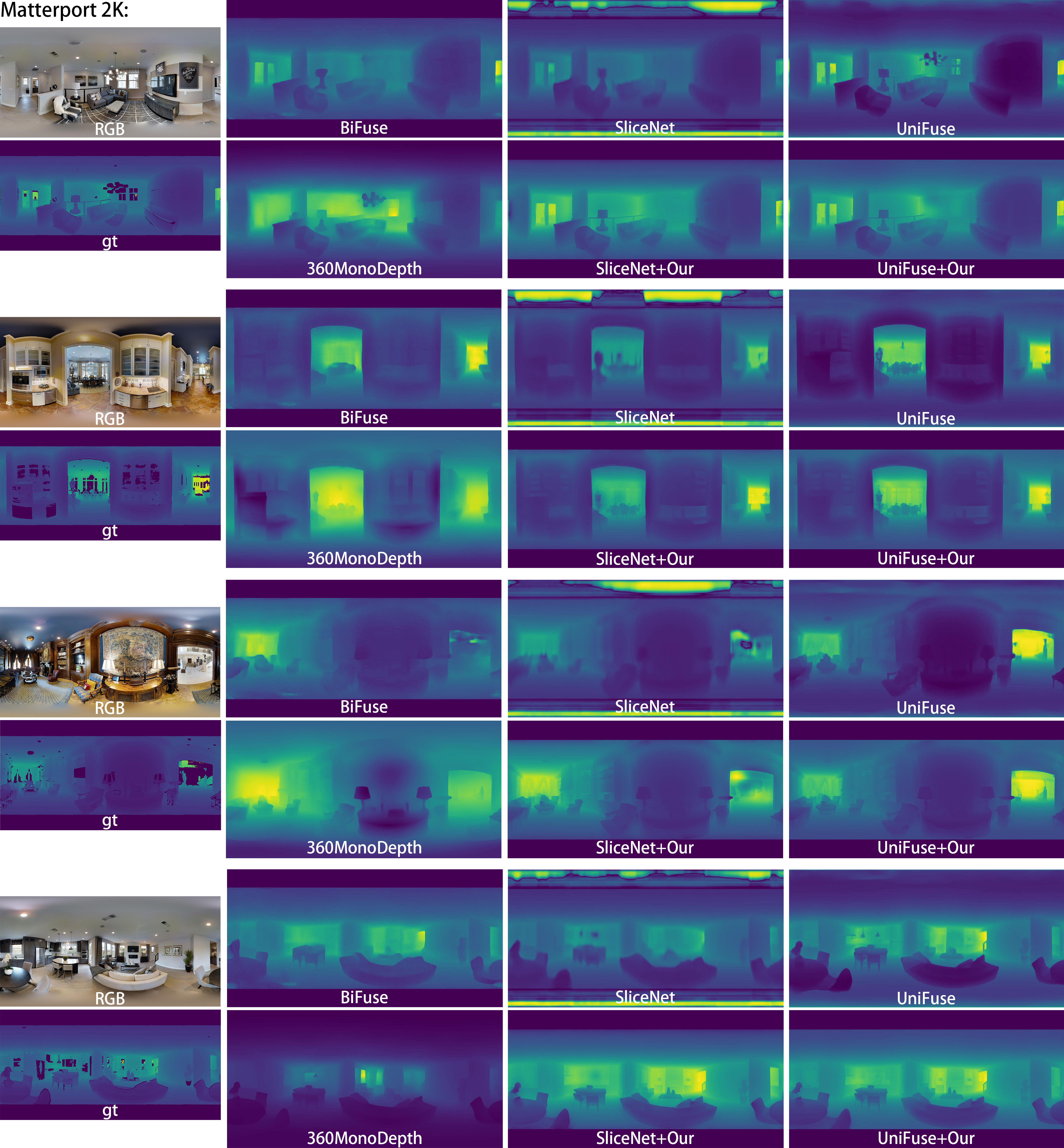}
  \caption{\label{fig:qualitative_matterport2}. Additional qualitative comparisons on the Matterport 2K dataset.} 
\end{figure*}

\begin{figure*}[t]
  \centering
  \includegraphics[width=1\linewidth]{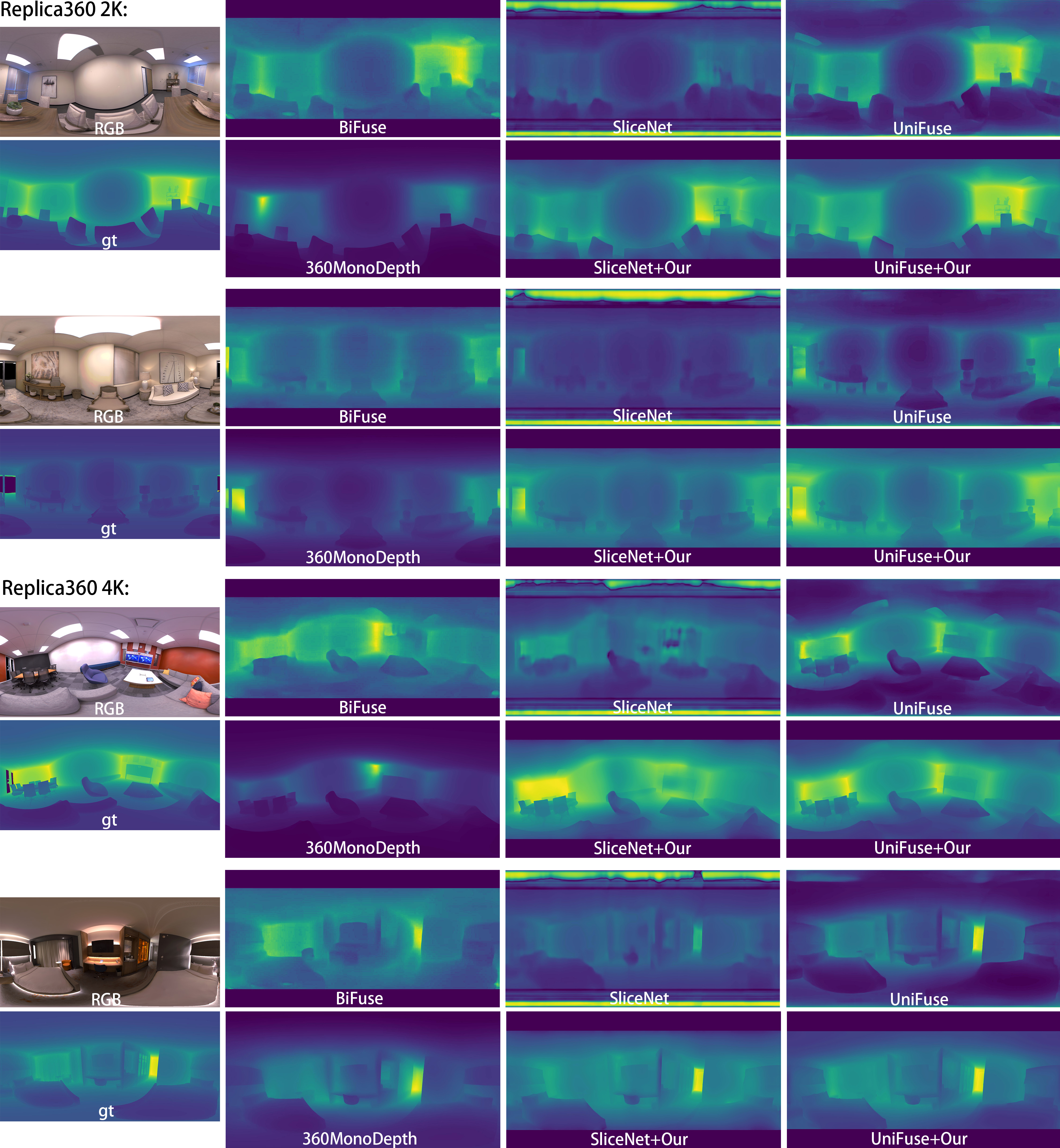}
  \caption{\label{fig:qualitative_replica}. Additional qualitative comparisons on the Replica 2K and 4K datasets.} 
\end{figure*}

We show additional qualitative results in Figure~\ref{fig:qualitative_matterport2} and Figure~\ref{fig:qualitative_replica}. In the first example of Replica360 4K result in Figure~\ref{fig:qualitative_replica}, we can see that the result of~\cite{reyarea2021360monodepth} has incorrect relative depths of the white wall w.r.t. the two adjacent walls.

\subsection{Quantitative metrics in 360MonoDepth~\cite{reyarea2021360monodepth}}

In~\cite{reyarea2021360monodepth}, they reported different quantitative scores on the Matterport 2K dataset and the Replica360 2K/4K datasets then our experiments. We think the differences came from the different approaches to align an estimated depth map and a ground-truth depth map. In our paper, we used the "median-scaling" approach that is the common practice in all panorama-based methods since OmniDepth~\cite{zioulis2018omnidepth}. After checking~\cite{reyarea2021360monodepth}'s paper and code, we think they instead used the least squares-based scaling approach (proposed in the original MiDaS v3 paper~\cite{9178977}), in which both optimal scaling and offset are computed by least-squares to align an an estimated depth map and a ground-truth one. We note that some of their quantitative scores are very different from the ones reported in other papers. For example, they reported RMSE of BiFuse as 0.994 on the Matterport 2K dataset, but it is 0.6259 on the Matterport 1K dataset reported in HoHoNet~\cite{sun2021hohonet}, SliceNet~\cite{pintoreslicenet}, UniFuse~\cite{jiang2021unifuse}, and OmniFusion~\cite{Li2022CVPR}. Our evaluation method reported RMSE of BiFuse as 0.6350 on the Matterport 2K dataset, which is much closer.


\subsection{Comparison with OmniFusion~\cite{Li2022CVPR}}

Recently, OmniFusion~\cite{Li2022CVPR} reported results with new state-of-the-arts quantitative scores among panorama-based methods. At submission, we could not get to run their codes correctly on our machines, so quantitative evaluations on the new Matterport 2K and Replica360 2K/4K datasets are not reported. Comparing to our method, OmniFusion outputs 1K instead of 2K or higher resolution outputs. We also don't find the same levels of details in their results (shown in their paper) as in stitching-based methods (\cite{reyarea2021360monodepth} and ours). We expect our method to benefit from the improved accuracy of panoramic depth maps generated by OmniFusion.

\subsection{Quantitative evaluation of sharp detail preservation}

We also measured several Laplacian-based metrics to quantitatively evaluate sharp details preservation. The results (Table 4) showed that our method significantly outperformed both panoramic methods and 360MonoDepth~\cite{reyarea2021360monodepth}. These quantitative results are in line with the qualitative comparisons shown in the main paper - both showed that our results outperform other panorama-based methods and 360MonoDepth (another stitching-based method) in terms of sharp detail preservation.

\begin{table}[h]
\small
\begin{center}
\begin{tabular}{| c | c | c c c |}
\hline
DS & Method & $\|\nabla^{2}\|$ $\downarrow$ & $\|LoG\|$ $\downarrow$ & \\
\hline
\parbox[t]{1.6mm}{\multirow{7}{*}{\rotatebox[origin=c]{90}{Matterport 2K}}} & Bifuse & 0.0450 & 0.2134 & \parbox[t]{1.6mm}{\multirow{7}{*}{\rotatebox[origin=c]{90}{Highlighting: \textcolor{red}{Best}, \textcolor{green}{second-best}, \textcolor{blue}{third-best} }}}\\
& SliceNet & 0.0386 & 0.1827 & \\
& UniFuse & 0.0402 & 0.1843 & \\
& 360MonoDepth & \textcolor{blue}{0.0384} & \textcolor{blue}{0.1681} & \\
&  Our (SliceNet) & \textcolor{green}{0.0287} & \textcolor{green}{0.1269} & \\ 
& Our (UniFuse) & \textcolor{red}{0.0281} & \textcolor{red}{0.1240} & \\
\hline

\parbox[t]{1.6mm}{\multirow{7}{*}{\rotatebox[origin=c]{90}{Replica360 2K}}} & Bifuse & 0.00093 & 0.0202 & \\
& SliceNet & 0.00397 & 0.0811 & \\
& UniFuse & 0.00134 & 0.0270 & \\
& 360MonoDepth & \textcolor{blue}{0.00064} & \textcolor{blue}{0.0122} & \\
& Our (SliceNet) & \textcolor{red}{0.00049} & \textcolor{red}{0.0107} & \\ 
& Our (UniFuse) & \textcolor{green}{0.00051} & \textcolor{green}{0.0110} & \\
\hline

\parbox[t]{1.6mm}{\multirow{7}{*}{\rotatebox[origin=c]{90}{Replica360 4K}}} & Bifuse & 0.00061 & 0.0132 & \\
& SliceNet & 0.00446 & 0.0918 & \\
& UniFuse & 0.00049 & 0.0103 & \\
& 360MonoDepth & \textcolor{blue}{0.00038} & \textcolor{blue}{0.0075} & \\
& Our (SliceNet) & \textcolor{red}{0.00024} & \textcolor{red}{0.0053} & \\ 
& Our (UniFuse) &  \textcolor{green}{0.00027} & \textcolor{green}{0.0058} & \\
\hline
\end{tabular}
\end{center}
\label{tab:comparisons_Laplacian}
\caption{$\|\nabla^{2}\|$ and $\|LoG\|$ measure the mean absolute errors of Laplacian and Laplacian of Gaussian (LoG) using the standard 5x5 mask, which are proxies to measure how sharp features of estimated and ground-truth depth maps match.}
\end{table}

\subsection{Using super-resolution approaches to up-sample panoramic depth maps instead of bilinear filtering}

We used a recent super-resolution method~\cite{Mei_2021_CVPR} to up-sample 1K outputs of UniFuse and SliceNet to 2K (Matterport3D) and 4K (Replica3604K). The results are very similar to bilinear up-sampling results qualitatively and quantitatively ($<1\%$ differences by RMSE, MAE, and AbsRel).

\end{document}